\documentclass[letterpaper, 10 pt, conference]{ieeeconf}  

\IEEEoverridecommandlockouts                              

\usepackage{graphics} 
\usepackage{epsfig} 
\usepackage{textcomp}
\usepackage{stfloats}
\usepackage{url}
\usepackage{verbatim}
\usepackage{graphicx}
\usepackage{cite}
\usepackage{tikz}
\usepackage{comment}
\usepackage{amsmath,amssymb} 
\usepackage{color}

\usepackage{wasysym}
\usepackage{graphics} 
\usepackage{epsfig} 
\usepackage{mathptmx} 
\usepackage{tikz}
\usepackage{comment}
\usepackage{color}

\usepackage{booktabs}
\usepackage{wasysym}
\usepackage{graphics} 
\usepackage{epsfig} 
\usepackage{mathptmx} 
\usepackage{color} 
\usepackage{tabularx}
\usepackage{makecell}
\usepackage{multicol}
\usepackage{mwe,lipsum}
\usepackage{array,multirow}
\usepackage{float}
\usepackage{eucal}
\usepackage{subcaption}
\usepackage{xcolor}
\usepackage{nicematrix}
\usepackage{color, colortbl}
\definecolor{LightCyan}{rgb}{0.88,1,1}
\usepackage{arydshln}

\usepackage[symbol]{footmisc}
\usepackage{pifont}

\newcommand{\cmark}{\ding{51}}%
\newcommand{\xmark}{\ding{55}}%

\usepackage{hyperref}

\usepackage[cal=cm]{mathalfa}

\title{\LARGE \bf
Event-Free Moving Object Segmentation from Moving Ego Vehicle
}

\author{Zhuyun Zhou$^{1,2}$, Zongwei Wu$^{2*}$, Danda Pani Paudel$^{3}$, Rémi Boutteau$^{5}$, Fan Yang$^{1}$, \\ Luc Van Gool$^{3,4}$, Radu Timofte$^{2}$, and Dominique Ginhac$^{1}$ 
\thanks{This research is financed in part by the French National Research Agency through ANR CERBERE (ANR-21-CE22-0006), by the Ministry of Education and Science of Bulgaria (support for INSAIT, part of the Bulgarian National Roadmap for Research Infrastructure), and by the Alexander von Humboldt Foundation.}
\thanks{$^{1}$Zhuyun Zhou, Fan Yang, and Dominique Ginhac are with University of Burgundy, Dijon, France. {\tt\small \{Zhuyun\_Zhou@etu., fanyang@, dginhac@\}u-bourgogne.fr }}%
\thanks{$^{2}$Zhuyun Zhou, Zongwei Wu, and Radu Timofte are with University of Wurzburg, Wurzburg, Germany. {\tt\small firstname.lastname@uni-wuerzburg.de}}
\thanks{$^{3}$Danda Pani Paudel and Luc Van Gool are with INSAIT, Sofia, Bulgaria. {\tt\small
firstname.lastname@insait.ai}}
\thanks{$^{4}$Luc Van Gool is also affiliated with CVL, ETH Zurich, Zurich, Switzerland. {\tt\small
vangool@vision.ee.ethz.ch}}
\thanks{$^{5}$Rémi Boutteau is with Université  Rouen Normandie, INSA Rouen Normandie, Université  Le Havre Normandie, Normandie Université, LITIS UR 4108, Rouen, France. {\tt\small remi.boutteau@univ-rouen.fr}}
\thanks{* Corresponding author}
}

\begin{document}

\maketitle
\thispagestyle{empty}
\pagestyle{empty}

\begin{abstract}

Moving object segmentation (MOS) in dynamic scenes is an important, challenging, but under-explored research topic for autonomous driving, especially for sequences obtained from moving ego vehicles. 
Most segmentation methods leverage motion cues obtained from optical flow maps. However, since these methods are often based on optical flows that are pre-computed from successive RGB frames, this neglects the temporal consideration of events occurring within the inter-frame, consequently constraining its ability to discern objects exhibiting relative staticity but genuinely in motion. To address these limitations, we propose to exploit event cameras for better video understanding, which provide rich motion cues without relying on optical flow. To foster research in this area, we first introduce a novel large-scale dataset called DSEC-MOS for moving object segmentation from moving ego vehicles, which is the first of its kind. For benchmarking, we select various mainstream methods and rigorously evaluate them on our dataset. Subsequently, we devise EmoFormer, a novel network able to exploit the event data. For this purpose, we fuse the event temporal prior with spatial semantic maps to distinguish genuinely moving objects from the static background, adding another level of dense supervision around our object of interest. Our proposed network relies only on event data for training but does not require event input during inference, making it directly comparable to frame-only methods in terms of efficiency and more widely usable in many application cases.
The exhaustive comparison highlights a significant performance improvement of our method over all other methods.
The source code and dataset are publicly available at: \href{https://github.com/ZZY-Zhou/DSEC-MOS}{https://github.com/ZZY-Zhou/DSEC-MOS}.

\end{abstract}

\section{INTRODUCTION}

Achieving precise moving object segmentations (MOS) in urban scenes plays a vital role in many vision tasks like image/video synthesis \cite{wu2020future,tancik2022block}. This segmentation is always achieved through cross-frame affinity matrices \cite{kundu2022panoptic,xie2022segmenting}. However, in more challenging scenarios, such as videos recorded from moving ego vehicles, conventional methods often falter, primarily due to the significant impact of unavoidable ego-motion \cite{philipp2019analytic,liang2023mpi}. Consequently, there exists a critical need for a high-performing MOS model from moving ego vehicles.

Recently, event cameras \cite{gallego2020event,guo2022low} have disrupted traditional computer vision paradigms. Different from frame-based cameras, event cameras operate asynchronously \cite{gallego2020event}, enabling them to capture rapid motion and react to dynamic scenes with remarkable precision. In addition, their sensitivity to pixel intensity changes and innate ability to adapt to varying illumination conditions make them a robust choice for challenging scenarios \cite{chen2020event,maqueda2018event}.

\begin{figure}[t]
  \centering
  
  \includegraphics[width=\linewidth]{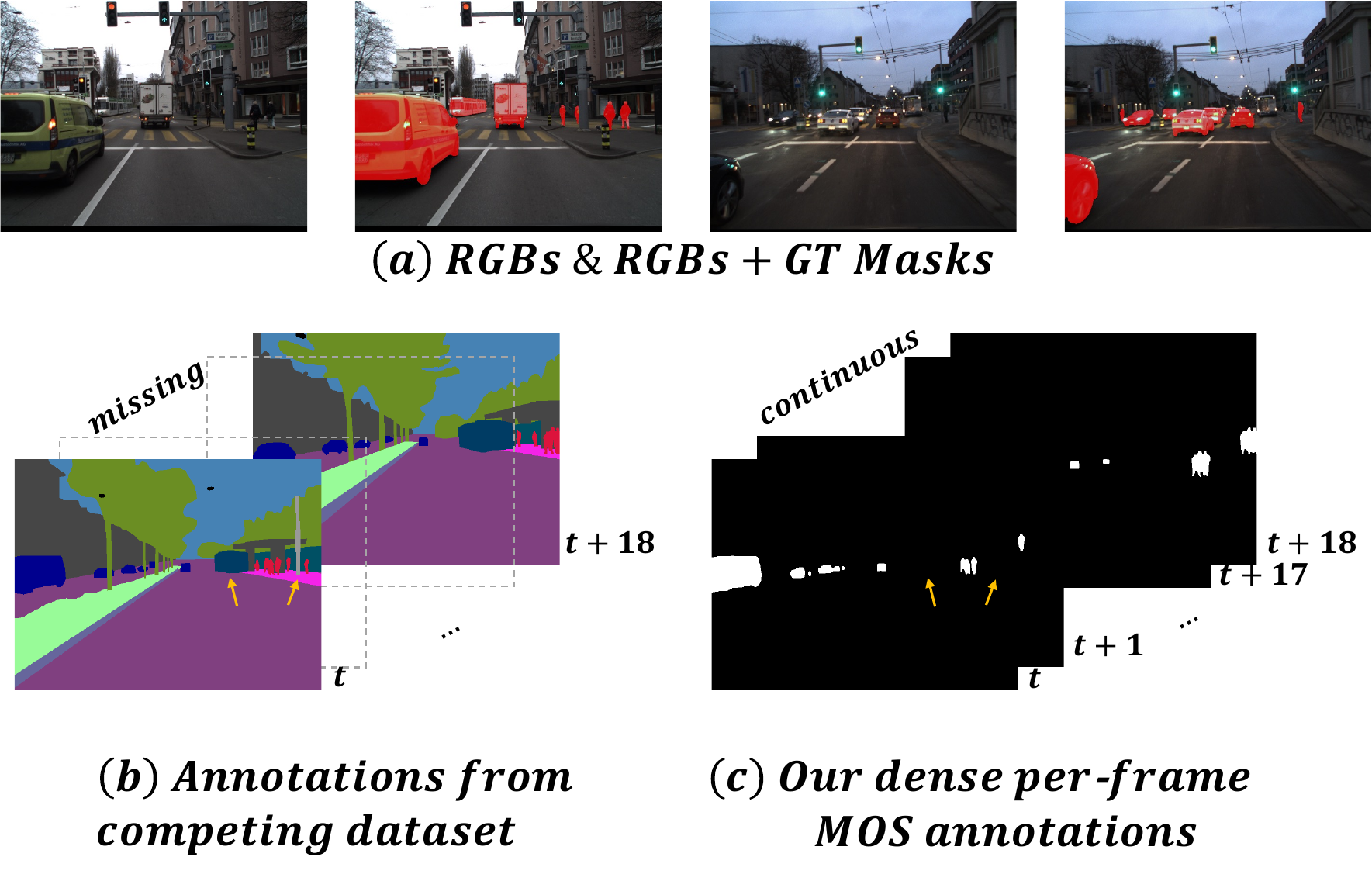}
  
  \vspace{-1mm}
  
  \caption{\textbf{DSEC-MOS Examples and Visual Comparison.} The top part (a) shows calibrated-to-Event RGB frames, and our DSEC-MOS Ground Truth Segmentation Masks visualized on calibrated RGB frames. The bottom part (b, c) shows that our dataset provides per-frame annotation and distinguishes the motion attributes, which are not available in the previous dataset \cite{xia2023cmda}. Best zoomed in.}
  
  \label{Fig-dataset}
  
  \vspace{-5mm}

\end{figure}

In this work, we aim to fully exploit the potential of event cameras for MOS from moving ego vehicles. Our problem extends the conventional (automatic) video object segmentation (VOS) \cite{zhou2022survey} in a broader, challenging, and more daily-usage oriented setting with three crucial differences: (a) VOS typically aims to segment salient or foreground objects \cite{hu2018unsupervised,pei2022hierarchical}, often a single object, while our goal is to segment all moving objects at once; (b) most automatic VOS methods preprocess the optical flow as input \cite{cho2023treating,zhang2021deep} while we leverage ground truth event data; (c) in VOS, the camera motion is usually negligible compared to the object motion, while in our scenarios both objects and the cameras (embedded in the ego vehicle) are in unknown motion. These differences make our task an exceptional challenge, which, to our knowledge, is the first time being addressed comprehensively.

To promote the research in the domain,
we introduce a substantial dataset of 13,314 frames, derived from the DSEC dataset \cite{gehrig2021dsec,zhou2023rgb}. This new dataset, called DSEC-MOS, covers a wide range of challenging driving scenarios, including different lighting conditions and complex traffic scenarios, as shown in Figure \ref{Fig-dataset}. It includes 8 categories of moving objects such as vehicles, pedestrians, cyclists, etc., each with different motion patterns, appearances, and scales, making the dataset highly suitable for both training and evaluation of MOS algorithms. Furthermore, DSEC-MOS provides dense pixel-level annotations for moving objects. 

Leveraging our proposed dataset, we performed a systematic benchmark involving mainstream object segmentation methods in video tasks. Intriguingly, we observe that the leading methods \cite{pei2022hierarchical,yuan2023isomer,karim2023med,cho2023treating} do not necessarily maintain their competitiveness on our challenging dataset. Nevertheless, with the same backbone, RGB-flow methods \cite{pei2022hierarchical,yuan2023isomer} consistently outperform RGB-only approaches \cite{cho2023treating,liu2021f2net}. Additionally, \cite{pei2022hierarchical,karim2023med} with deeper architectures such as transformers \cite{vaswani2017attention} outperform CNN-based counterparts \cite{yang2021learning,ren2021reciprocal} even when the latter uses extra flow cues. Consequently, we introduce a straightforward yet highly effective model, termed EmoFormer, that incorporates event clues as additional dense supervision during training while remaining event-free during inference.
Detailed comparative analyses show that our method significantly outperforms all existing SOTA approaches.

In summary, our contributions are 
three-fold:

\begin{itemize}
    \item We introduce a new MOS task with moving ego vehicles and present a generously sized and densely annotated dataset called DSEC-MOS. 
    \item To establish a benchmark, we rigorously select mainstream segmentation methods in video tasks with pure image input or with image-flow inputs.
    \item We introduce EmoFormer, which fully benefits from event clues during training while being event-free during inference. Our model outperforms all SOTA counterparts by a large margin and can serve as a strong baseline to facilitate further research in this domain.
\end{itemize}

\section{RELATED WORK}

\noindent\textbf{Automatic Video Object Segmentation (AVOS)}: AVOS represents a specialized task focused on segmenting objects within video sequences without prior knowledge about the target objects \cite{zhou2022survey}. Within the realm of AVOS, the primary cues for segmentation stem from the motion exhibited by objects. Traditional approaches involve computing similarity matrices \cite{wang2019learning,yang2021dystab,sun2022mining} across different video frames, implicitly harnessing temporal context to guide object identification. Another category of AVOS methods \cite{zhou2021flow,cho2022unsupervised,yang2019unsupervised,sultana2020unsupervised, zhou2020motion} explicitly uses the optical flow as a guiding cue. Considering that target objects in AVOS are typically salient and predominantly occupy the foreground, and camera motion is generally negligible in comparison to object motion \cite{pont20172017,perazzi2016benchmark,caelles20182018}, extracting the target object from the flow map is relatively straightforward compared to a real-world setting \cite{ding2023mose}. 

\noindent\textbf{Event Vision and Benchmarking:} Event cameras have attracted considerable research interest due to their unique ability to capture rapid motion through asynchronous event processing \cite{zuo2022devo,baldwin2022time}. Recent studies \cite{amini2022vista,ta2023l2e,ding2022spatio} have begun to explore the potential of event data for high-level tasks, especially in the context of driving scenarios. Pioneering works, such as \cite{gehrig2021dsec,zhu2018multivehicle}, have collected large RGB-Event datasets tailored to urban scenes. Nevertheless, these datasets often lack comprehensive manual high-level annotations. Some approaches resort to off-the-shelf techniques to generate pseudo ground truth annotations, such as the use of YOLO bounding boxes \cite{tomy2022fusing} or semantic segmentation \cite{sun2022ess}. However, the quality of such pseudo annotations may be limited. Furthermore, these efforts typically focus on global detection without distinguishing between motion features, i.e., discerning moving objects from static ones. The concurrent works \cite{mitrokhin2019ev,zhang2023multi,zhou2021event} detect moving objects but only a small number of sequences with indoor event input. In this paper, we introduce a novel large-scale MOS dataset that provides pixel-level annotations of scenes with moving ego vehicles. Our dataset stands as the first of its kind, designed specifically to tackle the distinctive challenges posed by moving objects in dynamic urban environments.

\begin{figure*}[t]
\centering

\includegraphics[width=.95\linewidth]{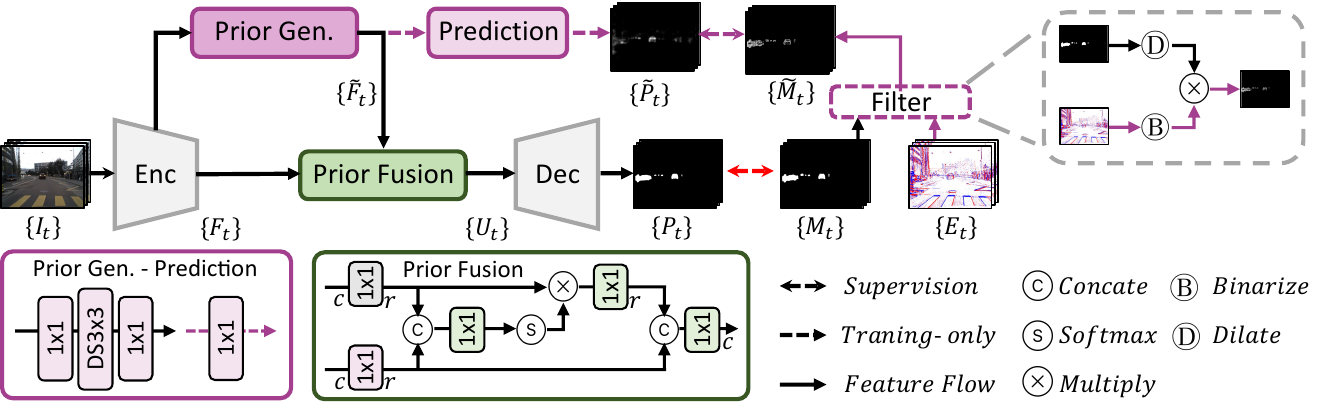}

\caption{\textbf{Architecture.} In addition to the standard RGB Encoder-Decoder architecture, we introduce an auxiliary branch dedicated to harnessing the motion insights derived from the recorded event data (Sec. \ref{generation}). This learned representation is subsequently merged into the main processing pipeline, thereby enhancing feature modeling (Sec. \ref{fusion}). To further refine event-based learning and the understanding of object dynamics, we employ semantic maps to transfer global scene motion into targeting objects' motion  (Sec. \ref{superv}). Such a 
filter 
strategy leads to a tightly coupled 
semantic-guided event 
awareness, ultimately shaping our joint learning scheme.}
\label{Fig-main}

\vspace{-2mm}

\end{figure*}

\section{METHODOLOGY}

 Given a video clip, denoted as $\{I_t \in \mathbb{R}^{3\times H \times W}\}^T_{t=1}$, comprising a sequence of $T$ frames, our objective is to predict masks $\{P_t \in \mathbb{R}^{H \times W}\}^T_{t=1}$ 
corresponding to the moving objects within the frames. These predicted masks are intended to closely align
with the ground truth masks $\{M_t \in \mathbb{R}^{H \times W}\}^T_{t=1}$ for the respective frames. 

In the following sections, we introduce our end-to-end learnable EmoFormer as shown in Figure \ref{Fig-main}. In contrast to the conventional approach of pre-computing the optical flow, our method harnesses motion cues directly from the recorded event data. Moreover, we seamlessly integrate the event prior as an additional dense supervision, resulting in event-free inference. 
This pivotal advance not only brings our model on par with RGB networks but also improves its suitability for real-world applications. For the sake of brevity in subsequent discussions, we omit the notion $\{._t\}^T_{t=1}$.

\subsection{Prior Generation}
\label{generation}

Without loss of generalizability, we take the encoded RGB feature $F
\in \mathbb{R}^{c \times h \times w}$ from the last layer of the backbone from one image as an example. While the RGB feature is rich in appearance information, it inherently lacks awareness of temporal variations. Although inner-frame motion can be modeled by comparing two consecutive images, the video's limited frames per second (FPS) restrict its ability to capture rapid motion. Hence, our objective is to estimate the rapid motion from a single RGB image, with the help of event data acquired during the frame time.

Specifically, we use a combination of 
depth-wise separable
convolutions as the generation module. First, we project $F
$ into a deeper latent feature space using a $Conv_{1\times 1}$ operation, capture local motion awareness with $DSConv_{3\times 3}$, and then project the result back using another $Conv_{1\times 1}$.  The choice of depth-wise separable convolution stems from the observation that rapid motion typically occurs within a localized region. Furthermore, the depth-wise convolution contributes to efficient processing. 

Consequently, we obtain an intermediate representation corresponding to the high dynamic features, denoted as 
$\Tilde{F} 
\in \mathbb{R}^{c \times h \times w}$. Mathematically, we have:
\begin{equation}
    \Tilde{F} = Conv_{1\times 1}(DSConv_{3\times 3}(Conv_{1\times 1}(F))) .
\end{equation}

To provide explicit guidance for the intermediate feature, we initially use a $Conv_{1\times 1}$ operation to generate a one-channel prediction map 
$\Tilde{P}$ 
based on 
$\Tilde{F}$. 
We then integrate event supervision, as described in Sec. \ref{superv}.

\subsection{Prior Fusion}
\label{fusion}

To seamlessly integrate the high dynamic feature 
$\Tilde{F}$ 
with the visual feature $F
$, we introduce a novel prior fusion framework shown in Figure \ref{Fig-main}. Traditional RGB-Event fusion methods such as \cite{tomy2022fusing} often operate in an equal or higher dimensional feature space. However, the direct application of such methods to our context poses a challenge. Since one feature is estimated based on the other, even with event supervision, it may still exhibit a high degree of redundancy compared to the other. Consequently, these methods may retain redundant information rather than mutually complementary ones.

We propose to project both of these features into a smaller subspace, denoted as $r \ll C$. This projection can be likened to a low-rank decomposition, wherein only the most informative modality-specific features are preserved. Moreover, this approach naturally reduces the computational overhead compared to alternative fusion techniques. Mathematically, we achieve this projection as follows:
\begin{equation}
    f = Conv_{1\times 1}(F) ; \quad \Tilde{f} = Conv_{1\times 1}(\Tilde{F}) .
\end{equation}

We then conduct cross-modal feature modeling within this lower-dimensional space $r$.

To facilitate this, we initially concatenate these feature maps, forming a correlation matrix $f_c\in \mathbb{R}^{r \times h \times w}$, followed by per-pixel attention using a global Softmax function:
\begin{equation}
\begin{split}
    Att = Softmax(\; f_c \;) ; \\
    \quad f_c = Conv_{1\times 1}\, ( \; Concat( \; f; \; \Tilde{f} \;)) .
\end{split}
\end{equation}

We then apply this attention map to the low-rank RGB feature, thereby mitigating the domain gap. 
Subsequently, 
we obtain the shared embedding by merging the attention-enhanced RGB feature and the motion feature by concatenation. Finally, this mixed embedding is projected back into the full-rank space $C$ to yield the fused feature $
U
\in \mathbb{R}^{c \times h \times w}$: 
\begin{equation}
    U = Conv_{1\times 1}\, ( \; Concat( \; f \times Att; \; \Tilde{f} \;)) .
\end{equation}

\subsection{Semantic-Guided Event Filter}

\label{superv}

Event data provide invaluable insights into the global scene motion. However, the dense distribution of events is not solely limited to areas occupied by moving objects, but can also encompass the contours of static or background elements. This complexity arises from the fact that our ego vehicle is also in motion, which makes it difficult to discern the motion exclusively associated with the target objects.

To solve this dilemma, we propose a novel approach that combines event data with ground truth semantic supervision. This hybrid strategy allows us to pass from the overall scene motion into the object motion, constraining the network attention around the moving object with dense supervision.

Specifically, it involves suppressing background regions by element-wise multiplication with the spatial semantic mask. However, the direct multiplication of the spatial semantic mask with the temporal event map can inadvertently remove essential motion cues, especially when the event motion map extends beyond the object boundaries due to motion blur. To mitigate this, 
We employ a dilation operation on the spatial mask with a specified kernel size before conducting 
semantic-guided event filter. 
Mathematical, let $M\in \mathbb{R}^{H \times W}$ be the ground truth (GT) semantic mask, $D\in \mathbb{R}^{3 \times 3}$ be the dilation matrix, and $E\in \mathbb{R}^{H \times W}$ be the event data. The 
filtered 
map $
\Tilde{M}
\in \mathbb{R}^{H \times W}$ is obtained by:
\begin{equation}
    \Tilde{M} = (M \oplus D) \circ E,
\end{equation}
where $\oplus$ is the morphological dilation and $\circ$ is the Hadamard product. In our application, we set $D$ as an all-ones matrix, and we treat positive and negative polarization changes equally in the binary form of event data.

Subsequently, we utilize 
$\Tilde{M}$ 
to supervise the intermediate representation 
$\Tilde{P}$ 
introduced in Sec. \ref{generation}. To minimize the disparity between these two maps, we leverage conventional MSE loss as the auxiliary loss denoted as 
$\Tilde{\mathcal{L}}$. 

\subsection{Overall Learning Pipeline}
Our event-guided training methodology departs from the conventional RGB learning pipeline by introducing additional semantic-guided event filter. On top of our RGB baseline \cite{karim2023med}, we incorporate the extra prior generation module (Sec. \ref{generation}) and the prior fusion module (Sec. \ref{fusion}). Our objective function can be summarized as follows:
\begin{equation}
 argmin \ (\; \mathcal{L}\, (\; \{P_t\}; \, \{M_t\} \;) + \Tilde{\mathcal{L}}\, (\; \{\Tilde{P}_t\}; \, \{\Tilde{M}_t\} \;) \;) .
\end{equation}
It is noteworthy that the event prediction $\{\Tilde{P}_t\}$ is exclusively employed during training to leverage event supervision, but it is not required during testing. During inference, our model operates solely on frames for outputting $\{P_t\}$ , ensuring an event-free process.

\section{Our DSEC-MOS Dataset}

\subsection{Dataset Annotation}
Our DSEC-MOS dataset builds upon the DSEC-MOD dataset \cite{zhou2023rgb}, originally designed for moving object detection with bounding box annotations. To create DSEC-MOS, we followed a systematic process:

\noindent\textbf{Candidate Mask Generation:} We initiated the process by employing the modified SOTA segmentation model \cite{kirillov2023segment} with bounding box prompts to generate candidate masks.

\noindent\textbf{Annotator Training:} Next, we train three annotators to become proficient in video segmentation tasks. They are provided with guidance and instructions on well-established VOS benchmarks to familiarize themselves with the task.

\noindent\textbf{Manual Verification:} After training, each annotator was tasked with reviewing the candidate masks, aided by reference to adjacent frames. If a candidate mask was found to be suboptimal with a visually incomplete or inaccurate segmentation, the annotator was asked to correct it manually.

\noindent\textbf{Specialization by Illumination Condition:} Since our dataset comprises sequences captured under various illumination conditions, we assigned each annotator to sequences from one typical condition to develop specializations.

\noindent\textbf{Cross and Double Verification:} The sequences from each illumination condition underwent cross and double verification. In case of uncertainty or discrepancies, annotators were encouraged to discuss and jointly modify the masks until a consensus was reached.

\begin{table}[t]
  \footnotesize

\begin{center}
  \caption{Datasets Per-Attribute Comparison}
    
    \vspace{-1mm}

\footnotesize

\setlength\tabcolsep{4pt}
\renewcommand{\arraystretch}{1.0}
\begin{tabular}{c c c c c c}

\toprule

Dataset & GT &  Outdoor & Per-Frame & MOS  \\

\midrule

EV-IMO \cite{mitrokhin2019ev} & \cmark & \xmark & - & \cmark  \\
DSEC-Sem \cite{sun2022ess} & \xmark & \cmark & \cmark & \xmark  \\ 
DSEC Night-Sem \cite{xia2023cmda} & \cmark & \cmark & \xmark & \xmark  \\

\midrule
\textbf{Ours} & \cmark & \cmark &  \cmark & \cmark  \\

\bottomrule

\label{tab:datasetcomp}

\end{tabular}
\end{center}

\vspace{-10mm}

\end{table}

\subsection{Dataset Comparison}

In Table \ref{tab:datasetcomp}, we offer a comprehensive per-attribute comparison with established event-based segmentation datasets. Significantly, our dataset emerges as a pioneering initiative, distinguished as the first of its kind to provide dense MOS masks from a moving ego vehicle. While EV-IMO \cite{mitrokhin2019ev} stands as another notable MOS dataset, it is confined to indoor settings and lacks paired RGB frames. This absence hinders its applicability for assessing RGB models, underscoring the unique nature of our dataset.

\begin{table*}[t]
\footnotesize
\centering
\caption{
\textbf{Quantitative Comparison} with SOTA AVOS Approaches on Our DSEC-MOS. Our EmoFormer outperforms both RGB and RGB-Flow counterparts with large margins. MS Inf. refers to Multi-Scale Inference.
Our EmoFormer achieves the best performances during both single-scale and multi-scale inferences.
}

\label{Tab-SOTA}

\begin{tabular}{l|c|c|c|c|c|c|c|c|c}

\toprule

Method & Pub. \& Year & Backbone & Optical Flow & MS Inf. & $\mathcal{J}$ Mean & $\mathcal{J}$ Recall & $\mathcal{F}$ Mean & $\mathcal{F}$ Recall & $\mathcal{J}$ \& $\mathcal{F}$ 
Mean
\\

\midrule

RTNet \cite{ren2021reciprocal} & CVPR'21 & ResNet-101 & \checkmark & - & 57.8 & 65.8 & 71.7 & 82.2 & 64.8 \\

AMCNet \cite{yang2021learning} & ICCV'21 & ResNet-101 & \checkmark & - & 60.6 & 71.3 & 76.1 & 86.8 & 68.4 \\
TMO \cite{cho2023treating} & WACV'23 & ResNet-101 & \checkmark & - & 56.2 & 62.1 & 73.0 & 82.8 & 64.6 \\
Isomer \cite{yuan2023isomer} & ICCV'23 & Swin-T & \checkmark & - & 55.9 & 65.3 & 72.2 & 86.0 & 64.1 \\

\hdashline

F2Net \cite{liu2021f2net} & AAAI'21 & ResNet-50 & - & - & 45.1 & 48.4 & 66.9 & 74.5 & 56.0 \\
TMO \cite{cho2023treating} & WACV'23 & ResNet-101 & - & - & 53.5 & 57.1 & 70.3 & 81.8 & 61.9 \\
MED-VT \cite{karim2023med} & CVPR'23 & Swin-B & - & - & 58.0 & 67.3 & 81.0 & 90.6 & 69.5 \\
\rowcolor[RGB]{238, 218, 242}

\textbf{EmoFormer} & \textbf{Ours} & Swin-B & - & - & \textbf{66.5} & \textbf{72.9} & \textbf{82.2} & \textbf{92.2} & \textbf{74.4} \\

\midrule

HFAN \cite{pei2022hierarchical} & ECCV'22 & Swin-T & \checkmark & \checkmark & 62.6 & 69.5 & 76.8 & 85.7 & 69.7 \\

MED-VT \cite{karim2023med} & CVPR'23 & Swin-B & - & \checkmark & 63.5 & 71.8 & 81.5 & 90.3 & 72.5 \\

\rowcolor[RGB]{238, 218, 242}  

\textbf{EmoFormer} & \textbf{Ours} & Swin-B & - & \checkmark & \textbf{68.9} & \textbf{77.7} & \textbf{83.7} & \textbf{93.1} & \textbf{76.3} \\

\bottomrule

\end{tabular}

\vspace{-2mm}

\end{table*}

\section{EXPERIMENTS}

\subsection{Experimental Setup}
\noindent\textbf{Implementation Details:} We take video-swin \cite{liu2022video} as the encoder that extracts features from 6 frames. Following \cite{karim2023med}, each input frame is resized to 360$\times$409. Similar to \cite{zhou2023rgb}, we leverage event data from 50ms, in accordance with the frame time. Classic data augmentation methods such as multi-scale training and flipping are adopted. We adopt AdamW as the optimizer, with an initial learning rate 1e-5 and 1e-4 for the backbone and others, respectively. The weight decay is set to 1e-4 and the polynomial learning rate decay is employed with a power of 0.9. Our network is built upon Pytorch. The whole network training takes around 2 days on a 3090 GPU. Unlike most VOS works, our method does not require any pretraining on large-scale datasets. For benchmarking, we retrain all the SOTA counterparts from their official resources. For practical application consideration, we exclude the time-burden CRF post-processing for all.

\noindent\textbf{Evaluation Metrics:} We use classical evaluation metrics, including region similarity $\mathcal{J}$ and contour accuracy $\mathcal{F}$, in accordance with the VOS tasks \cite{perazzi2016benchmark}. We compute both the mean and recall 
measures 
for each metric, as well as the overall performance $\mathcal{J} \& \mathcal{F}$ based on 
their mean measures.

\subsection{Benchmark and Comparison}

It is noteworthy that our benchmark does not include comparisons with RGB-Event object segmentation methods, as none 
have been identified thus far. Despite the availability of certain semantic segmentation techniques \cite{jia2023event,xia2023cmda,sun2022ess,zhang2021issafe}, 
these methods do not deal with the distinction between moving and static objects, so they do not fall within the scope of our benchmark.

\begin{figure}[t]

  \centering

  \includegraphics[width=.9\linewidth]{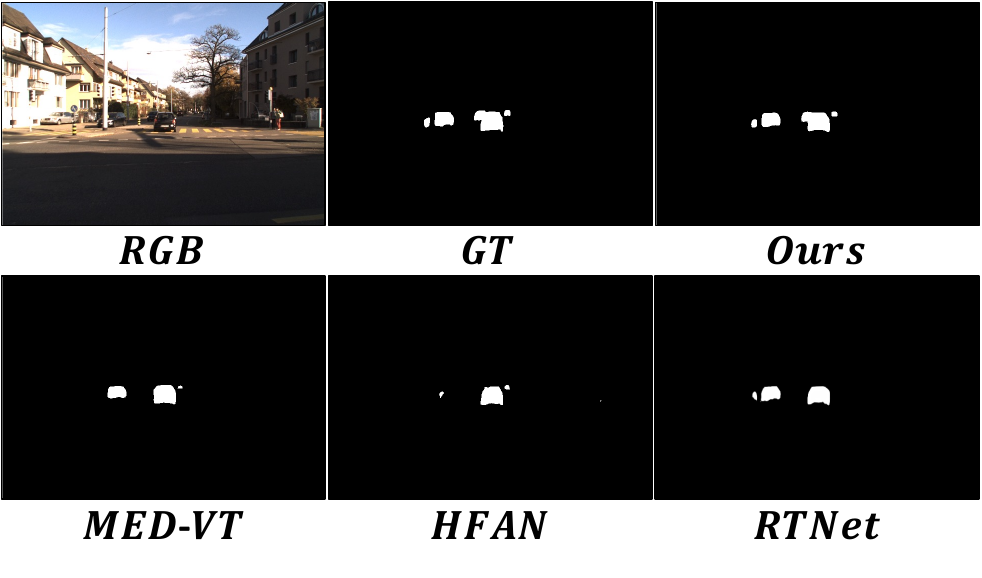}
  
  \vspace{-3mm}
  
  \caption{\textbf{Qualitative Comparison.} Our generated masks are closer to the 
  Ground Truth (GT) compared to the counterparts.
  Please zoom in for details.
  }
  \label{Fig-quali}
    
  \vspace{-5mm}
    
\end{figure}

\noindent\textbf{Comparison against SOTA Methods:}
Our moving object segmentation task is very similar to 
Automatic 
Video Object Segmentation (AVOS) tasks; in that we do not provide manual clues such as initial masks or bounding boxes to identify the target objects. Instead, we aim to automatically segment moving objects based solely on video sequences. To evaluate the effectiveness of our method, we compared 8 state-of-the-art automatic VOS methods as reported in Table \ref{Tab-SOTA}. To ensure fair comparisons, we evaluated our performance with both single-scale and multi-scale inference techniques. The results demonstrate a significant performance improvement over all other methods. In the case of single-scale inference, our network significantly improves the performance of our RGB baseline (MED-VT) with an absolute gain of +8.5\% $\mathcal{J}$ Mean, outperforming the current state-of-the-art method, AMCNet \cite{yang2021learning}. 
It is worth noting that AMCNet requires both RGB and flow maps during inference, whereas our method operates only on RGB frames. In the case of multi-scale inference, we achieve an absolute gain of +5.4\% $\mathcal{J}$ Mean over MED-VT. We also provide qualitative comparisons in Figure \ref{Fig-quali}, illustrating that our method excels at exploring temporal cues and generating prediction masks that closely match ground truth, outperforming any pure RGB or RGB-Flow counterparts.

\subsection{Ablation Studies}

\begin{table}[t]
\footnotesize
\centering
\caption{
\textbf{Ablation Study on Supervision Strategies.}
$\circ$ is the Hadamard multiplication to suppress static responses.
}

\vspace{-1mm}

\label{Tab-Guide}

\begin{tabular}{l|c|c|c|c|c|c}

\hline

\hline

\multicolumn{2}{c|}{Aux. Sup. Source} & $ \circ$ GT & Dilation & $\mathcal{J}$ M & $\mathcal{F}$ M & $\mathcal{J}$ \& $\mathcal{F}$ 
M
\\

\hline

\hline

1 & - & - & - & 63.5 & 81.5 & 72.5 \\
2 & Flow & - & - & 66.2 & 80.7 & 73.5 \\
3 & Semantic & - & - & 67.2 & 82.2 & 74.7 \\

4 & Semantic & - & \checkmark & 67.6 & 83.1 & 75.4 \\
5 & Event & - & - & 65.4 & 80.1 & 72.8 \\
6 & Event & \checkmark & - & 67.9 & 82.8 & 75.4 \\

\rowcolor[RGB]{238, 218, 242}  

7 & Event & \checkmark & \checkmark & \textbf{68.9} & \textbf{83.7} & \textbf{76.3} \\

\hline

\hline

\end{tabular}

\end{table}

We perform multi-scale inference for ablation studies on Supervision Strategies.
In our training pipeline, we leverage event data as auxiliary supervision, facilitated by a semantic mask. To assess the effectiveness of this approach, we conducted experiments in which we replaced the intermediate representation with alternative supervision methods, such as optical flow only, semantic mask only, and event data only. The quantitative results are shown in Table \ref{Tab-Guide}. It is evident that, while the auxiliary supervisions consistently improve our baseline, their individual performance is inferior to our full approach. We observed that omitting the semantic map guidance and the dilation operation leads to performance deterioration. This outcome can be attributed to the role of semantic guidance in attenuating noisy event responses from static regions. In addition, dilation plays a crucial role in the alignment of spatial masks with temporal motion blur.

\section{CONCLUSION}

We demonstrate the effectiveness of using the event prior for moving object segmentation in dynamic scenes. Our method combines the inherent advantages of event data for capturing global motion with semantic supervision. This leads to an intermediate representation that enables the network to separate moving objects from the static back- ground, resulting in another level of dense supervision. Moreover, our method no longer requires any event input during inference, which makes it particularly efficient and suitable for real-world applications based on standard RGB cameras. We also introduce a new large-scale dataset designed for moving object segmentation in dynamic autonomous driving scenarios, which is the first of its kind. Our method significantly outperforms the state of the art and confirms the potential of event-based pipelines for autonomous driving. To sum up, we hope that our method and dataset can promote further research in this field.

\bibliographystyle{IEEEtran}
\bibliography{IEEEexample}

\end{document}